\documentclass[sigconf]{acmart}
\usepackage{todonotes} 
\usepackage{framed}
\usepackage{multirow}
\usepackage{algorithm}
\usepackage{algorithmicx}
\usepackage{algpseudocode}
\usepackage{xcolor}
\usepackage{subfigure}
\usepackage{footnote}
\usepackage{enumerate}

\newcommand{\db}[1]{[\kern-0.15em[ #1
 ]\kern-0.15em]}

\newcommand{\method}{DeepFAIT }
\newcommand{\methode}{DeepFAIT}

\newcommand{\questions}{3 }

\usepackage{amsfonts}

\usepackage{pifont}

\begin{document}
\title{Fairness Testing of Deep Image Classification with Adequacy Metrics}

\author{Peixin Zhang}
\affiliation{%
  \institution{Zhejiang University}
}

\author{Jingyi Wang}
\affiliation{%
  \institution{Zhejiang University}
}

\author{Jun Sun}
\affiliation{%
  \institution{Singapore Management University}
}

\author{Xinyu Wang}
\affiliation{%
  \institution{Zhejiang University}
}

\begin{abstract}
As deep image classification applications, e.g., face recognition, become increasingly prevalent in our daily lives, their fairness issues raise more and more concern. It is thus crucial to comprehensively test the fairness of these applications before deployment. Existing fairness testing methods suffer from the following limitations: 1) applicability, i.e., they are only applicable for structured data or text without handling the high-dimensional and abstract domain sampling in the semantic level for image classification applications; 2) functionality, i.e., they generate unfair samples without providing testing criterion to characterize the model's fairness adequacy. To fill the gap, we propose \methode, a systematic fairness testing framework specifically designed for deep image classification applications. \method consists of several important components enabling effective fairness testing of deep image classification applications: 1) a neuron selection strategy to identify the fairness-related neurons; 2) a set of multi-granularity adequacy metrics to evaluate the model's fairness; 3) a test selection algorithm for fixing the fairness issues efficiently. We have conducted experiments on widely adopted large-scale face recognition applications, i.e., VGGFace and FairFace. The experimental results confirm that our approach can effectively identify the fairness-related neurons, characterize the model's fairness, and select the most valuable test cases to mitigate the model's fairness issues.  

\end{abstract}

\maketitle

\section{Introduction}
\label{sec:introduction}
Deep learning (DL) has created a new programming paradigm in solving many real-world problems, e.g., computer vision~\cite{face_recognition}, medical diagnosis~\cite{medical_diagnosis}, and natural language processing~\cite{nlp}. 
However, DL is far from being trustworthy to be applied in certain ethic-critical scenarios, e.g., toxic language detection~\cite{toxic} and facial recognition~\cite{racial_aug}, as decisions of DL models can be unfair, i.e., discriminating minorities or vulnerable subpopulations, which has raised wide public concern~\cite{trust_ai}. 
Therefore, just like traditional software, it is only more demanding to test the fairness of DL models systematically before their deployment.

Unfortunately, there still lacks a commonly agreed definition on fairness. Existing fairness formalization typically concerns different sub-populations~\cite{demographic,verification,fairness,counterfactual}. These sub-populations are normally determined by different domains (values) of sensitive attributes (e.g., race and gender), which are often application-dependent. To name a few, demographic parity defines that minority candidates should be classified at an approximately same rate as majority members~\cite{demographic,verification}. Individual discrimination states that a well-trained model must output the approximately same predictions for instances (i.e., pairs of instances) which only differ in sensitive attributes~\cite{fairness,counterfactual}. We refer the readers to~\cite{science} for detailed definitions of fairness and remark that we focus on individual fairness in this work.

Multiple recent works~\cite{themis,aequitas,sg,adf} have investigated the fairness\footnote{So far restricted to individual fairness.} testing problem of machine learning models. For instance, THEMIS~\cite{themis} first aims to measure the frequency of unfair samples by randomly sampling the value domain of each attribute. AEQUITAS~\cite{aequitas} integrates a global and local phase to search for unfair samples in the input space more systematically.
Symbolic Generation~\cite{sg} utilizes a constraint solver~\cite{concolic} to solve the path on the local explanation decision tree~\cite{lime} of a given seed sample to acquire a large number of diverse unfair samples. State-of-the-art work ADF~\cite{adf,tse} and its variants~\cite{issta21} adopt a gradient-guided search strategy to identify unfair samples more effectively. Despite the considerable progress, existing fairness works still suffer from the following limitations: 1) applicability, i.e., they are only applicable for structured data or text without handling the high-dimensional and abstract domain sampling in the semantic level for image classification applications; 2) functionality, i.e., they generate unfair samples without providing testing criterion to characterize the model's fairness adequacy.  
\begin{figure*}[t]
\centering
\includegraphics[width=0.8\textwidth]{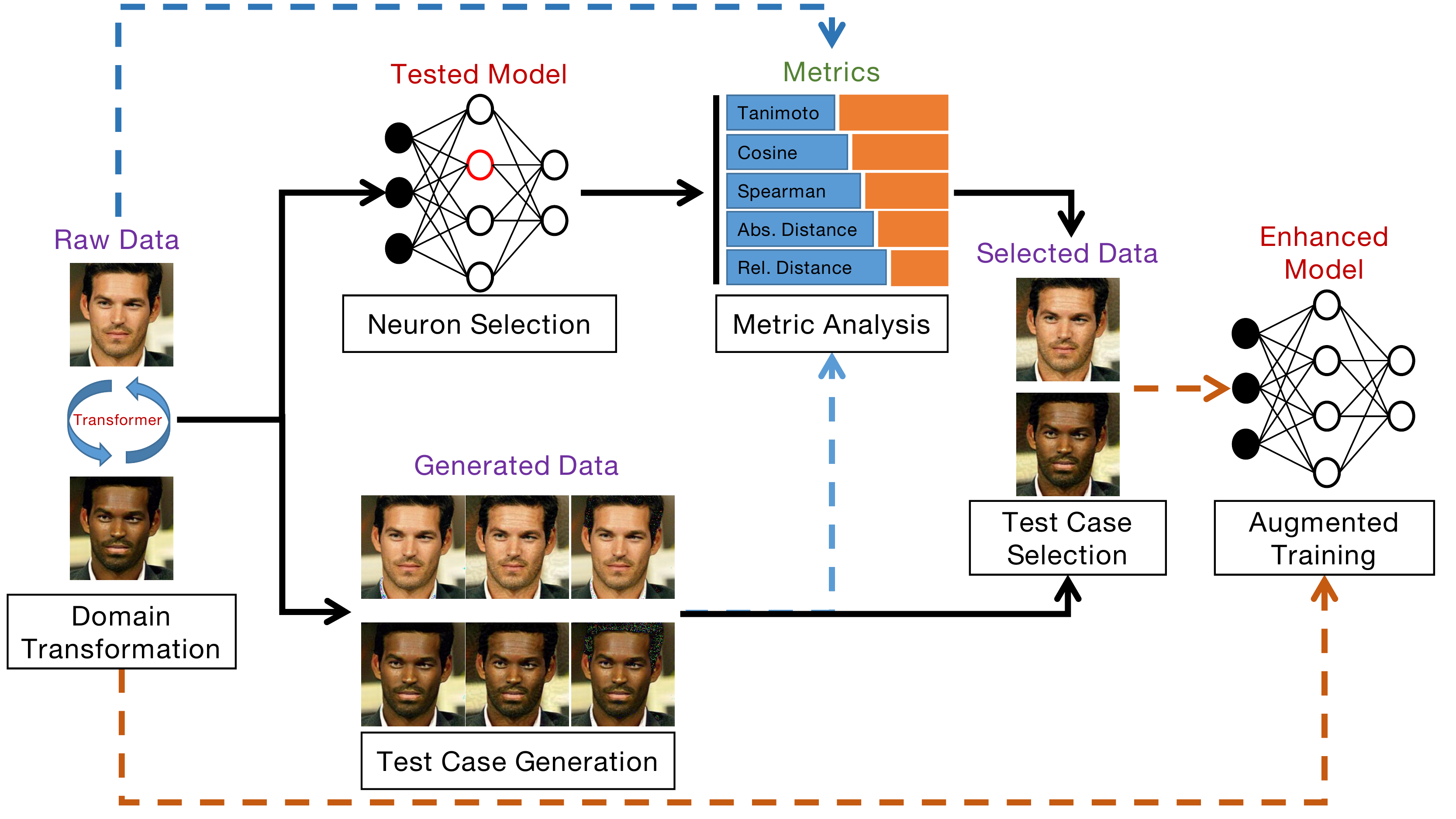}
\caption{Overview of \methode . Among the generated data, from left to right, the image pairs are crafted by Random Generation (RG), Gradient-based Generation (GG), and Gaussian-noise Injection (GI), respectively. The blue and brown dashed lines indicate the process of measuring testing adequacy and fairness enhancement, respectively.}
\label{fig:overview}
\end{figure*}

To fill the gap, we propose a systematic fairness testing framework named \methode, which is especially designed for evaluating and improving the fairness adequacy of deep image classification applications. \method provides several key functionalities enabling effective fairness testing of image applications: 1) a neuron selection strategy to identify the fairness-related neurons; 2) a set of multi-granularity adequacy metrics to evaluate the model's fairness; 3) a test selection algorithm for fixing the fairness issues efficiently. We address multiple technical challenges to realize \methode. Specifically, as shown in Fig. \ref{fig:overview}, \method consists of five modules. First, we adopt a widely-used image-to-image transformation technology, i.e., Generative Adversarial Network (GAN)~\cite{cyclegan}, to transform images across the sensitive domains. Then, we apply significance testing on the activation differences of neurons to obtain those fairness-related neurons and design 5 testing metrics both on layer- and neuron-level based on the identified fairness-related neurons. Next, we implement three test case generation strategies including fairness testing method and image processing processing technology to generate a variety of unfair samples. Last, we propose a test selection algorithm to select more valuable test case to repair the model with smaller cost. 

\method has been implemented as an open-source self-contained toolkit. We have evaluated \method on widely adopted large-scale face recognition datasets (VGGFace~\cite{vggface} and Fairface~\cite{fairface}). The results show that compared with DeepImportance~\cite{importance_driven}, \method is more capable of identifying fairness-sensitive neurons of the model. Furthermore, the proposed testing metrics calculated on these neurons are highly correlated with fairness and can be used to guide the search of unfair samples effectively. More importantly, the fairness issues can be fixed by selecting a small amount of test cases with our test selection algorithm to further train the model.

In a nutshell, we make the following technical contributions:
\begin{itemize}
	\item We propose a systematic fairness testing framework specially designed for deep image classification applications consisting a set of multi-granularity fairness adequacy metrics on fairness-related neurons.
	\item Based on the proposed adequacy metrics, we propose a test selection algorithm to evaluate the value of each test case in improving the model's fairness to reduce the cost of fixing the model.
	\item We implemented \method as a self-contained toolkit, which can be freely accessed online\footnote{https://github.com/icse44/DeepFAIT}.	The evaluation shows that the proposed testing criteria in \method are well correlated with the fairness of DL models and is effective to guide the selection of unfair samples. 
\end{itemize}

We frame the reminder of the paper as follows. We provide the necessary background on DNN and robustness testing criteria in Section~\ref{sec:preliminary}. We then present \method in detail in Section~\ref{sec:methodology}. In Section~\ref{sec:experiment}, we discuss the experimental setup and results. Lastly, we briefly review the releted works in Section~\ref{sec:relatedwork} and conclude our work in Section~\ref{sec:conclusion}.

\section{Preliminaries}
\label{sec:preliminary}

\subsection{Deep Neural Network}
\label{subsec:dnn}


In this work, we focus on the fairness testing of deep learning models, specifically, deep neural networks (DNNs) for image classification applications. 
A deep neural network $F$ consists of multiple hidden layers between an input and an output layer. It can be denoted as a tuple $M=(I, L, \Phi, TS)$ where
\begin{itemize}
\item $I$ is the input layer;
\item $L = \{L_{j}|j \in \{1, \dots, J\}\}$ is a set of hidden layers and the output layer, the number of neurons in the $j$-th layer is $|L_j|$, and the $k$-th neuron in layer $L_{j}$ is denoted as $n_{j,k}$ and its output value with respect to the input x is $v_{j,k}(x)$; 
\item $\Phi$ is a set of activation functions, e.g., Sigmoid, Hyperbolic Tangent (TanH), or Rectified Linear Unit (ReLU);    
\item $TS$: $L \times \Phi \to L$ is a set of transitions between layers. As shown in Equation~\ref{eq:neuron}, the neuron activation value, $v_{j,k}(x)$, is computed by applying the activation function to the weighted sum of the activation value of all the neurons within its previous layer, and the weights represent the strength of the connections between two linked neurons.
\begin{equation}
\label{eq:neuron}
v_{j,k}(x) =\phi(\sum_{l=1}^{|L_{j-1}|} \omega_{j-1,k,l} \cdot v_{j-1, l}(x))
\end{equation}
\end{itemize}
 
A classification DNN can be defined as $M:X \to Y$ which transforms a given input $x\in X$ to an output label $y\in Y$ by propagating layer by layer as above.

\subsection{Individual Fairness for Image Classification}
\label{subsec:definition}
We denote the fairness sensitive attribute of interest as $SA$ (e.g., race). Note that for image classification, $SA$ is \emph{hidden} from $X$. We define $HF:X\to SA$ as a function which returns the sensitive attribute of a given sample $x\in X$. We further define $X_A\subset X$ as the samples satisfying $HF(x)=A$ where $x\in X, A\in SA$. 
To change the sensitive attribute for a sample $x$, we define a transformation function $T_{A \to B}:X\to X$ which transforms a sample from $X_A$ to $X_B$ while preserving other information. Then, we define individual fairness of an image classification model $M$ as follows~\cite{science}.
\begin{definition}[Individual Fairness]
\label{de:if}
Given an image classification model $M$ trained on $X$, we define that it is individually fair iff there exists no data $x \in X$ satisfying the following conditions:
\begin{itemize}
	\item $x \in X_A, x'=T_{A \to B}(x)$
	\item $M(x)\neq M(x')$.
\end{itemize}
On the other hand, $x$ (and $x'$) is called an unfair sample if $x$ satisfies the above conditions.
\end{definition}

\subsection{Robustness Testing Criteria}
\label{subsec:robust testing}

A variety of robustness testing criteria for DNN has been proposed~\cite{deepxplore,deepgauge,surprise,robot,deepgini,importance_driven}. We briefly introduce the following representative robustness testing metrics. Readers are referred to~\cite{deepxplore,deepgauge,surprise,robot,deepgini,importance_driven} for details.

\vspace{1mm}
\noindent\textbf{Neuron Activation} Neuron coverage~\cite{deepxplore} is the first robustness testing metric for DNN, which computes the percentage of activated neurons, i.e., neuron values greater than a threshold. Later, DeepGauge~\cite{deepgauge} extends it with multi-granularity neuron coverage criteria from two different levels: 1) neuron-level, e.g., k-Multisection Neuron Coverage, Neuron Boundary Coverage and Strong Neuron Activation Coverage, focusing on the value distribution of a single neuron, and 2) layer-level to measure the ranking of the neuron values in each layer, e.g., Top-k Neuron Coverage and Top-k Neuron Patterns. Surprise Adequacy~\cite{surprise} evaluates the similarity between test case and training data based on the kernel density estimation or Euclidean distance of neuron activation traces. Importance-driven coverage~\cite{importance_driven} measures the value of neurons from another perspective, i.e., the contribution of each neuron within the same layer with respect to the prediction.

\vspace{1mm}
\noindent\textbf{Output Impurity} Unlike the aforementioned work, DeepGini~\cite{deepgini} only takes the output vector into consideration, which measure the likelihood of misclassification by Gini impurity.

\vspace{1mm}
\noindent\textbf{Loss Convergence} RobOT~\cite{robot} proposed First-Order Loss to measure the convergence quality inspired by the observation that if we perturb the instance based on its gradient with respect to the loss, the loss will increase and gradually converge~\cite{resistant}.

\subsection{Problem Definition}
Different from the previous robustness testing works, we aim to propose a set of testing adequacy metrics for individual fairness specially designed for image classification. In particular, we aim to achieve the following research objectives: 
\begin{itemize}
    \item How can we design testing adequacy metrics which are well correlated with the model's fairness?
    \item How can we select test cases which can effectively fix the model's fairness issues? 
\end{itemize} 

\section{DeepFAIT Framework}
\label{sec:methodology}

As shown in Figure~\ref{fig:overview}, \method systematically tests, evaluates and improves a DNN's fairness with the following 4 main components:
\begin{enumerate}[1)]
\item \textit{Domain transformation.} We develop a method based on CycleGAN~\cite{cyclegan} to realize the transformation function $T_{A\to B}$. 
\item \textit{Fairness-related neuron selection.} We propose to conduct testing in a more effective way by filtering out neurons which are strongly correlated with the model's fairness.  

\item \textit{Multi-granularity metric analysis.} We design a set of multi-granularity fairness testing coverage metrics to measure the adequacy of fairness testing. These metrics are particularly calculated based on the selected fairness-related neurons to be more effective. 

\item \textit{Fairness enhancement.} We develop a set of test case generation algorithms to identify a diverse set of unfair samples for mitigating discrimination by augmented training on these unfair samples. To further reduce the cost, we also propose a test selection algorithm to select more valuable test cases for the model's fairness enhancement based on the proposed metrics.
\end{enumerate}
In the following, we present the details of each component.

\subsection{Domain Transformation}
\label{subsec: transfer}
\begin{figure}[t]
\centering
\includegraphics[width=0.4\textwidth]{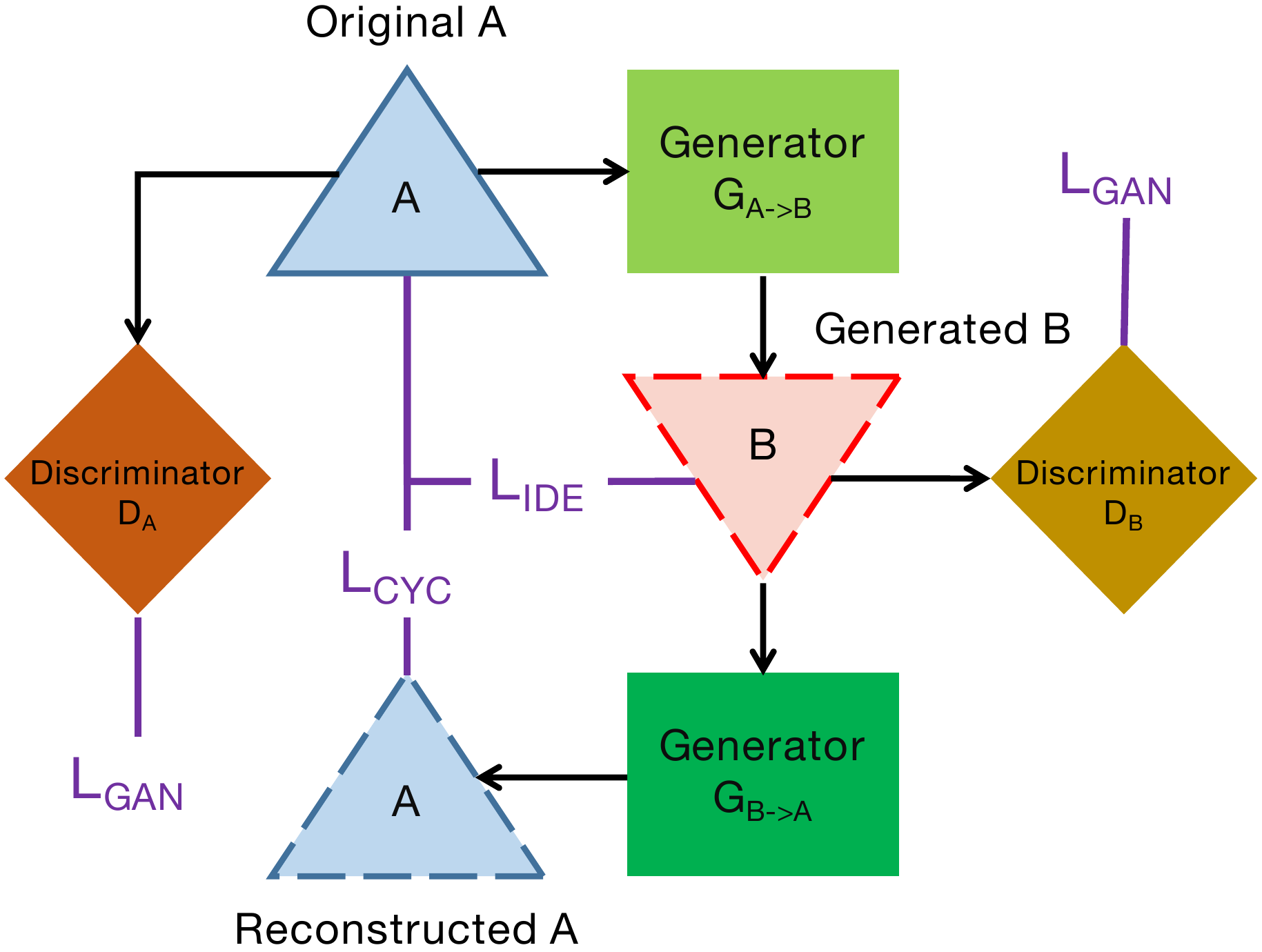}
\caption{Structure of CycleGAN model (A to B). The blue and red triangles indicate the data A and B respectively, and the solid line and dotted frame represent the original and generated data respectively. The black arrow indicates the data flow and the purple line indicates the calculation of loss function. The models for B to A has similar structure.}
\label{fig:cyclegan}
\end{figure}

The first question is how to realize the domain transformation function $T_{A\to B}$.
Note that this is straightforward for structured or text data which can be done by replacing the protected feature or token with a value from a predefined domain~\cite{adf,tse}.
However, for image data, the sensitive attribute of interest is hidden from the input feature space. 
We thus follow ~\cite{racial_aug} and adopt CycleGAN~\cite{cyclegan} to transform images across different protected domains as follows.

As shown in Figure~\ref{fig:cyclegan}, CycleGAN provides a mechanism to transfer from $A$ to $B$ domains and $B$ to $A$ domains respectively with two corresponding generative models $T_{A \to B}, T_{B \to A}$. Similar with traditional GAN~\cite{gan}, it contains two discriminators $D_A$ and $D_B$ to distinguish whether the input is `real', i.e., a sample is generated by generative model or from the original dataset.

The loss function consists of three parts, the first one is adversarial loss, which is defined as follows, 
\begin{equation}
\begin{split}
\label{eq:gan}
L_{GAN}(T_{A \to B}, D_B, A, B) &= \mathbb{E}[\log(1-D_B(T_{A \to B}(X_A)))] \\
&+ \mathbb{E}[\log D_B(X_B)]
\end{split}
\end{equation}
The transformer $T_{A \to B}$ aims to synthesize a picture satisfying the distribution of $X_B$ based on the seed from $X_A$, while the goal of discriminator $D_B$ is to distinguish the raw images $X_B$ from the artificial ones $T_{A \to B}(X_A)$. $T_{B \to A}$ and $D_A$ have the same definition of adversarial loss. 
Since domain transformation needs to modify other information as little as possible in the process of changing the sensitive attribute, the second part of the objective function is thus to ensure that the generated image is identical with the raw one, which is defined as follows:
\begin{equation}
\begin{split}
\label{eq:ide}
L_{IDE}(T_{A \to B}) &= \mathbb{E}[\|T_{A \to B}(X_A)-X_A\|_p]
\end{split}
\end{equation}
In addition, it also needs to ensure that $T_{A \to B}(X_A)$ is in the data distribution of $X_B$. To this end, CycleGAN introduces cycle consistency loss as the core of its joint optimization objective to make the synthesized image more realistic. It is defined as follows.
\begin{equation}
\begin{split}
\label{eq:cyc}
L_{CYC}(T_{A \to B}, T_{B \to A}) &= \mathbb{E}[\|T_{B \to A}(T_{A \to B}(X_A))\|_p] \\ 
&+ \mathbb{E}[\|T_{A \to B}(T_{B \to A}(X_B))\|_p]
\end{split}
\end{equation}
The intuition is that the pair of well-trained generators can recover the original image through the reconstruction process, i.e., for a given $x \in X_A$, $T_{B \to A}(T_{A \to B}(x)) = x$. 
Overall, the complete loss function of CycleGAN is defined as follows.
\begin{equation}
\begin{split}
\label{eq:loss}
L &= L_{GAN}(T_{A \to B}, D_B, A, B) + L_{GAN}(T_{B \to A}, D_A, B, A)\\
&+ \gamma(L_{IDE}(T_{A \to B})+L_{IDE}(T_{B \to A})) \\
&+ \eta L_{CYC}(T_{A \to B}, T_{B \to A})
\end{split}
\end{equation}
where $\gamma$ and $\eta$ are hyperparameters to balance among these three losses. In the ``Raw Data'' frame of Figure~\ref{fig:overview}, we show an example race transformer (Caucasian to African) from VGGFace.

\subsection{Fairness-related Neuron Selection}
\label{subsec: neuron}

The next step is to select the fairness-related neurons, i.e., those neurons in the DNN which may have a significant impact on the fairness of the model's decisions. The benefit is that, rather than blindly selecting neurons in certain layers \cite{deepgauge} or only the model's output \cite{deepgini}, it enables us to test the model's fairness in a more fine-grained and focused way as these neurons are more correlated with the model's fairness. 

Our key idea of neuron selection is by significance testing \cite{kw} of whether a neuron is fairness-related as follows.
Formally, we make the following null hypothesis on a given neuron $n_{j,k}$.
\begin{equation}
\label{eq:hy}
\begin{split}
&H_0 = n_{j,k} \ \text{is} \ \text{not} \ \text{fairness-related}\\
\end{split}
\end{equation}
In addition, we use a standard parameter (a.k.a., significance level), $\alpha$, to control the probability of making errors, e.g., rejecting $H_0$ when $H_0$ is true.
We then use Kruskal-Wallis test~\cite{kw} (also know as H-test) to test the above hypothesis. The intuition is that we could identify a fairness-related neuron by looking at the difference between the activation distribution on those fair samples and unfair samples. The larger the difference, the more fairness-related is the neuron. Specifically, given the training dataset $X$ and $T$, we could collect the activation value difference on neuron $n_{j,k}$ over $X$:
\begin{equation}
\label{eq:diff}
XD = \{\|v_{j,k}(x)-v_{j,k}(x')\| | \forall x \in X, x'=T(x)\}.
\end{equation}
We further divide $XD$ into two orthogonal subsets $XD_f, f=0,1$, according to the prediction results $M(x)$ and $M(x')$ are equal or not, respectively.

We first sort $XD$ in ascending order and denote the rank of i-th element in $XD_f$ as $r_f^i$. Then we add the ranks in each subset $XD_f$ to obtain the rank sum, denoted as $R_f = \sum_i r_f^i$. When the original hypothesis $H_0$ is true, the average rank of each subset should be close to that of all samples, i.e., $(|XD|+1) / 2$, and we thus use the following equation to compute the rank variance between subsets:
\begin{equation}
\label{eq:rvs}
RVS = \sum_{f=0,1}|XD_f|(\frac{R_f}{|XD_f|}-\frac{|XD|+1}{2})^2
\end{equation}
In order to eliminate the influence of dimension, we then calculate the average of rank variance of all samples, which is defined as follows.
\begin{equation}
\begin{split}
\label{eq:arv}
ARV &= \frac{1}{|XD|-1}\sum_{f=0,1}\sum_{i=1}^{|XD_f|}({r_f^i}-\frac{|XD|+1}{2})^2\\
&=\frac{|XD|(|XD|+1)}{12}
\end{split}
\end{equation}
Note that the freedom degree of sample variance is $|XD|-1$. Thus, the Kruskal-Wallis rank-sum statistic H is given by,
\begin{equation}
\label{eq:kw}
H = \frac{RVS}{ARV} = \frac{12}{|XD|(|XD|+1)}\sum_{f=0,1}\frac{R_f^2}{|XD_f|}-3(|XD|+1)
\end{equation}
When $|XD|$ is large, $H$ approximately obeys a chi-square distribution~\cite{chi_square} with the freedom degree of $1$, $H \sim  \mathcal{X}^2(1)$. Therefore, the critical value of Kruskal-Wallis testing, $H_c$, corresponding to $\alpha$ is determined according to the chi-square distribution table. That is, if the computed statistic $H>H_c$, we reject $H_0$ and conclude that the neuron $n_{j,k}$ is fairness-related.

\subsection{Fairness Adequacy Metrics}
\label{subsec:metric}
Next, we propose a set of testing metrics to measure the adequacy of DNN fairness. Note that different from the robustness testing metrics~\cite{deepgauge,importance_driven,robot,deepgini,surprise}, fairness testing metrics are based on the behavioral differences between an instance pair, i.e., $x$ and $x'$ (which only differ in certain sensitive attributes). Note that one important desirable property on the metrics is that they should be well correlated with the model's fairness.

Our fairness metrics will satisfy the above property as we define our metrics on the basis of selected fairness-related neurons. Specifically, let $NF_j \subset \{n_{j,1}, \dots, n_{j,|L_j|}\}$ be a set of fairness-related neurons within layer $L_j$. We denote the activation value vector and boolean activation patterns over neurons in $NF_j$ with respect to the input $x$ as $\vec{v}(x,NF_j)$ and $\vec{a}(x,NF_j)$, respectively. Note that in $\vec{a}(x,NF_j)$, $1$ and $0$  represent whether the value of $n_{j,k} \in NF_j$ before and after ReLU is the same or not, respectively. We first characterize the model differences between the inputs $x$ and $x'$ at the layer level as the decision is propagated layer by layer. 

\vspace{1mm}
\noindent\textbf{Tanimoto Coefficient} Tanimoto coefficient~\cite{tanimoto} is a similarity ratio defined over bitmaps. In DNN, the activation of a neuron indicates whether the abstract features of the neuron will be used in the subsequent decision-making process. Then as shown in Equation~\ref{eq:tan}, we compute the division of the number of common activated neurons (i.e., nonzero bits) over the number of neurons activated by either sample.
\begin{equation}
\label{eq:tan}
TC(x,x',NF_j) = \frac{\vec{A}\cdot \vec{A'}}{\|\vec{A}\| + \|\vec{A'}\|-\vec{A}\cdot \vec{A'}}
\end{equation}
where $\vec{A} = \vec{a}(x,NF_j)$ and $\vec{A'} = \vec{a}(x',NF_j)$.

\vspace{1mm}
\noindent\textbf{Cosine Similarity} Cosine similarity is one of the most commonly used similarity measure in machine learning applications. For example, it is often used to judge whether two faces belong to the same person in face recognition~\cite{lfw,racial_aug}. We calculate the cosine similarity of the activation traces in the layer representation space, which is defined as follows: 
\begin{equation}
\label{eq:cos}
CS(x,x',NF_j) = \frac{\vec{v}(x,NF_j)\cdot \vec{v}(x',NF_j)}{\|\vec{v}(x,NF_j)\|\cdot\|\vec{v}(x',NF_j)\|}
\end{equation}

\vspace{1mm}
\noindent\textbf{Spearman Correlation} Spearman correlation~\cite{spearman} is a non-parametric statistic to measure the dependence between the rankings of two variables. Although it discards some information, i.e., the real activation value, it retains the order of neurons' activation status, which is an essential characteristic, since the neurons within the same layer often learn similar features and the closer to the output layer an activated neuron is, the more important it is for the model's decision~\cite{deepgauge}. Formally, spearman correlation is computed by
\begin{equation}
\label{eq:spe}
SC(x,x',NF_j) = 1 - \frac{6\sum_{k=1}^{|NF_j|}{(r(\vec{v}_{j,k}(x))-r(\vec{v}_{j,k}(x')))^2}}{|NF_j|(|NF_j|^2-1)}
\end{equation}
where $r(\cdot)$ is the rank function.

The above layer-level metrics are based on the statistical results on a set of neurons in a layer. 
In addition, we also provide a more fine-grained metric based on a single neuron.

\vspace{1mm}
\noindent\textbf{Neuron Distance} As presented in~\ref{subsec:dnn}, each neuron within the same layer independently learns and extracts features from its previous layer. Therefore, neuron distance is a fine-grained metric to characterize the diversity of model behavior differences. In particular, for each fairness-related neuron $n_{j,k} \in |NF_j|$, the distance between $x$ and $x'$ is denoted as $nd(x, x',n_{j,k})$, which is defined as follows:
\begin{equation}
\label{eq:distance}
nd(x,x',n_{j,k}) = \left\{
\begin{aligned}
|v_{j,k}(x)-v_{j,k}(x')|, & &absolute\\
\frac{\max(v_{j,k}(x), v_{j,k}(x'))}{\min(v_{j,k}(x), v_{j,k}(x'))}, & &relative
\end{aligned}
\right.
\end{equation}
Note that the absolute and relative distance are only computed when $n_{j,k}$ are activated by both $x$ and $x'$~\cite{cover}.

Finally, 
we could define the coverage for each testing metric by dividing its value range (based on $X$ and the domain transformer $T$) into $Z$ equal bins, denoted as $B^z$, and then compute the coverage ratio as follows:
\begin{equation}
\label{eq:coverage}
\frac{\sum_{i=1}^{|C|}|\{B_i^z | \exists x \in X: C_i(x,T(x)) \in B_i^z\}|}{Z*|C|}
\end{equation}
where $|C|$ is the dimension for collecting the metric value. For layer-level similarity, we traverse layer by layer to calculate its value, thus $|C|=|J|$, and the value range is $[0,1]$ for Tanimoto and Cosine coefficient and $[-1,1]$ for Spearman correlation. For neuron distance, we select $topK$ neurons each layer to analyse, thus $|C|=|J|*topK$, and the upper bound is computed by the training data~\cite{deepgauge}. 

\subsection{Fairness Enhancement}
\label{subsec: generation}
The above metrics enables us to evaluate a model's fairness adequacy. 
The follow-up questions are 1) how to generate diverse test cases to improve the fairness adequacy, and then 2) how to select the most valuable test cases to enhance the model's fairness by augmented training, which has been proved to be useful in enhancing the robustness or fairness
of DNN~\cite{deepgauge,adf,tse}.

We address the first question by introducing the following test case generation methods, both from the spirit of fairness testing literature and traditional digital image processing approaches. Given a fair sample pair $(x,x')$, where $x$ and $x'$ are only different in certain sensitive attributes, we aim to maximize the output differences between them after applying perturbation $p$ on them, which is formally defined as follows:
\begin{equation}
\label{eq:objective}
\text{argmax}_p \{M(x+p)-M(x'+p)\}
\end{equation}
We introduce the following perturbation methods for approximating the above goal.

\vspace{1mm}
\noindent\textbf{Random Generation (RG)} Random testing is the most common testing method in software testing, and is adopted by THEMIS~\cite{themis} and AEQUITAS~\cite{aequitas}. Here, we select the perturbed pixel and perturbing direction randomly as follows,
\begin{equation}
\label{eq:RG}
p = random(-1, 0, 1) * step\_size,
\end{equation}
where $0$ means the pixel value is retained, $1$ and $-1$ represent increasing and decreasing the pixel value respectively multiplied by a $step\_size$. 

\vspace{1mm}
\noindent\textbf{Gradient-based Generation (GG)} It is noticeable that gradient is an effective tool to generate test cases for DL model~\cite{adf,tse,fgsm,jsma}. We utilize the gradient-based method in~\cite{adf,tse} to approximate the solution of the optimization problem~\ref{eq:objective}, which relies on the sign of gradient of loss function with respect to the input, i.e., $sg = sign(\bigtriangledown_x J(x, y))$ and $sg' = sign(\bigtriangledown_x' J(x', y))$. As shown in Equation~\ref{eq:GG}, we then choose the pixel with the same direction (sign) and the corresponding direction for perturbation.
\begin{equation}
\label{eq:GG}
p =  (sg_i==sg_i') * sg_i * step\_size
\end{equation}

\vspace{1mm}
\noindent\textbf{Gaussian-noise Injection (GI)} Gaussian noise is a commonly used noise in image domain (especially for RGB image), which is physically caused by poor lighting and high temperature of sensors. Therefore, the natural synthetic images are often acquired through applying Gaussian perturbations~\cite{gaussian1, gaussian2,importance_driven}. The probability density of Gaussian noise (perturbation) $p$ is defined as follows:
\begin{equation}
\label{eq:GI}
f(p) = \frac{1}{\sqrt{2\pi}\sigma}e^{-\frac{(p-\mu)^2}{2\sigma^2}}
\end{equation}
where $\mu$ and $\sigma$ are the mean and standard deviation respectively.


After generating test cases to increase the testing adequacy, i.e., the coverage of metrics, we then address the second question which selects a subset of test cases for augmented training to improve the model's fairness more efficiently. 
In particular, we prefer to select a set of test cases with diverse metric values inspired by \cite{robot}. 
Note that the neuron distance for each sample is a vector, of which each value is collected from each neuron independently which is difficult to quantify accurately. We thus only utilize the layer-level metrics, i.e., Tanimoto coefficient, cosine similarity, and Spearman correlation, for the selection.

Specifically, we adopt the K-Multisection Strategy (KM-ST) in our work, which uniformly sample the value space of a given metric. Formally, let $C_{min}$ and $C_{max}$ be the minimum and maximum value of crition $C$ collected from all the test cases, $n$ is the total number of test cases we need, and $k$ is the number of divided sections. We first divide the value range $[C_{min}, C_{max}]$ into $k$ equal sections with the interval of $d = (C_{max}-C_{min}) / k$, and then randomly select $n/k$ samples from each section independently to consist the final augmenting dataset.

\section{Experiments}
\label{sec:experiment}
We have implemented \method as a self-contained toolkit based on Pytorch. We have published the source code, along with all the experiment details and results online. The evaluations are conducted on a server with 1 Intel Xeon 3.50GHz CPU, 64GB system memory, and 2 NVIDIA GTX 1080Ti GPU.

\subsection{Experimental Setup}
\label{subsec:setup}
\textbf{Datasets and Models} We adopt $4$ open-source datasets in our evaluation. Two of them are used for training the domain transformer (refer to Section~\ref{subsec: transfer}) of the protected attribute \textit{gender} and \textit{race} respectively, and the remaining two are the experiment subjects.
\begin{itemize}
	\item \emph{CelebA~\cite{celeba,celeba_link}} is a widely used large-scale face recognition dataset. It contains $202,599$ face images from $10,177$ celebrities around the world. Each image has $5$ landmark locations and $40$ binary attributes, e.g., male, big lips, and pale skin. We use CelebA for gender transformation.
	\item \emph{BUPT-Transferface~\cite{bupt_transferface,bupt_link}} is a dataset of face images which aims to bring racial awareness into studies on face discrimination and achieving algorithm fairness. 
	We use it for race transformation.
	\item \emph{VGGFace~\cite{vggface,vggface_link}} consists of over $2,600,000$ images for $2,622$ identities from IMDB. The dataset is collected from multiple search engines, e.g., Google and Bing, with limited manual curation. We evaluate \method on VGGFace. 
	\item \emph{FairFace~\cite{fairface,fairface_link}} contains $108,501$ face images collected from the YFCC100m dataset.	It is manually labeled with \emph{gender}, \emph{race}, and \emph{age} groups and balanced on \emph{race}. It divides the age into $9$ bins. We train an age classifier and take it as another benchmark for evaluating \methode. 
\end{itemize}

We utilize two state-of-the-art DNN architectures as the benchmark models.
\begin{itemize}
	\item \emph{VGG~\cite{vgg}} is an advanced architecture for extracting abstract features from image data. VGG utilizes small convolution kernels (e.g., $3*3$ or $1*1$) and pooling kernels (e.g., $2*2$) to significantly increase the expressive power of the model.
	\item \emph{ResNet~\cite{resnet}} improves traditional sequential DNNs by solving the vanishing gradients problem when expanding the number of layers. It utilizes short-cuts (also called skip connections), which adds up the input and output of a layer and then transforms the sum to the next layer as input.
\end{itemize}
The details of our pre-trained models are shown in Table~\ref{tab:acc}.

\begin{table}[t]
\centering
\caption{Accuracies of the experimented models.}
\label{tab:acc}
\begin{tabular}{|c|c|c|}
\hline
Dataset & Model & Accuracy \\
\hline
VGGFace & VGG-16 & 97.28\% \\ \hline
FairFace & ResNet-50 & 96.08\% \\
\hline
\end{tabular}
\end{table}

\vspace{1mm}
\noindent\textbf{Face Annotation and Transfer}
For VGGFace, we crawl the racial and gender information by retrieving the celebrity idenetities from an active fan community FamousFix.com~\cite{annotation}. Then we check manually to make sure the meta information is correct. For Fairface, the metadata about sensitive attributes is downloaded with the images.

In the training process of the transformation model, we need to ensure that only the given sensitive attribute changes to avoid the impact of other sensitive attributes. Therefore, when we train the race (respectively gender) transformation model, the image pairs that we construct are made sure to have the same gender (respectively race).

\vspace{1mm}
\noindent\textbf{Parameters}
Table~\ref{tab:configuration} shows the values of all the parameters used in our experiment to run \methode, which either follows the settings of existing approaches or are decided empirically.

\begin{table}[t]
\centering
\caption{Configuration of experiments.}
\label{tab:configuration}
\begin{tabular}{|c|c|c|}
\hline
Parameter & Value & Description\\
\hline
$\gamma$ & 5 & the weight of $L_{IDE}$ in CycleGAN \\
$\eta$ & 10 & the weight of $L_{CYC}$ in CycleGAN \\
$\alpha$ & 0.05 & the significance level\\
step\_size & 5 & the step size of pertubation in RG and GG\\
$\mu$ & 0 & the mean of gaussian distribution in GI\\
$\sigma$ & 7 & the std of gaussian distribution in GI\\
$maxIt$ & 10 & the maximum iteration of generation\\
$topK$ & 10 & the top-$k$ fairness-sensitive neurons\\
$Z$ & 1000 & the number of bins\\
\hline
\end{tabular}
\end{table}

\subsection{Research Questions}
\label{subsec:rq}
We aim to answer the following \questions research questions through our evaluation.

\vspace{1mm}
\noindent\textbf{RQ1: Is \method effective in identifying fairness-related neurons?}
\begin{figure*}[t]
\subfigure[\method (Race)]{
\includegraphics[width=0.235\textwidth]{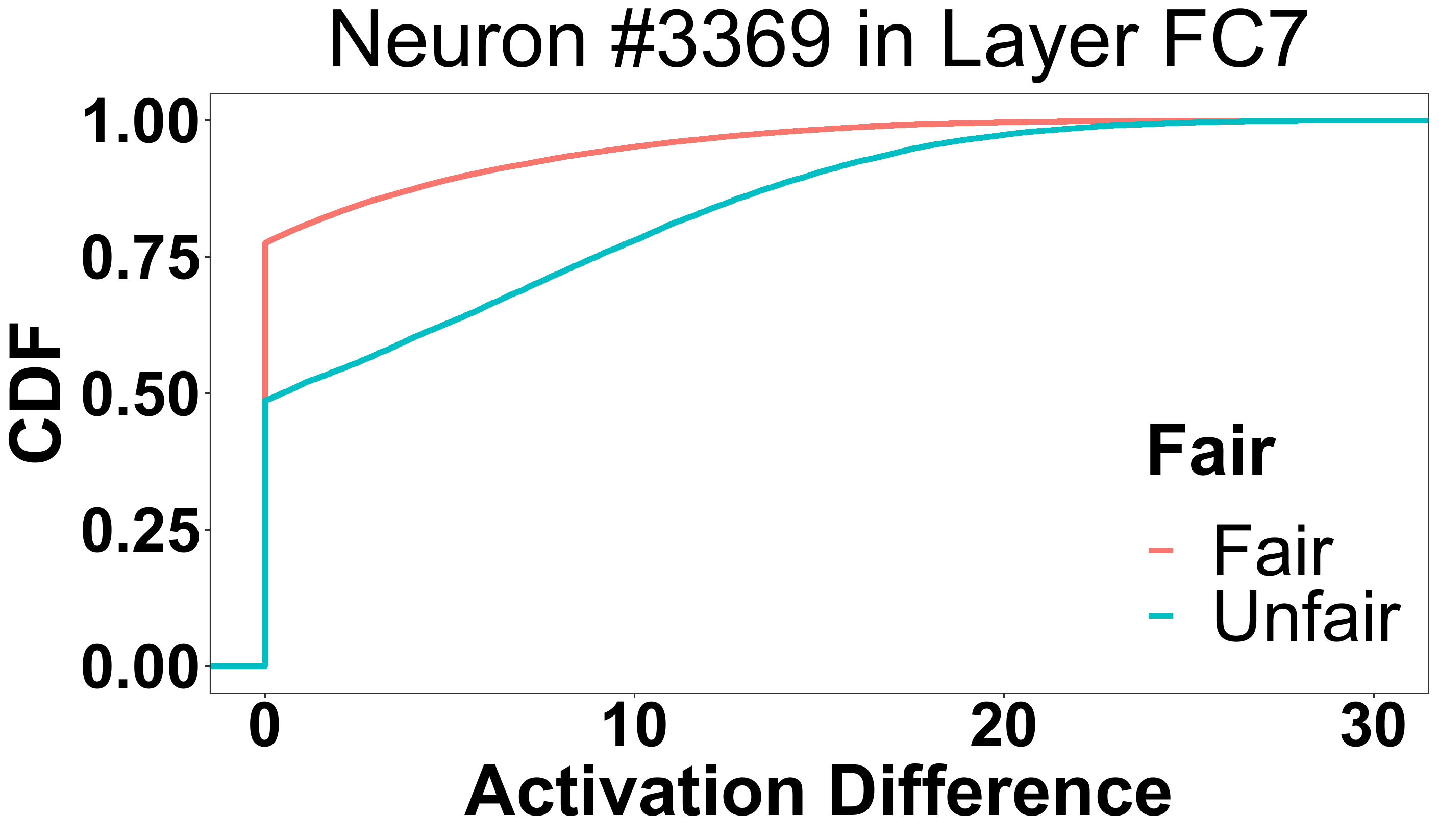}
\label{fig:race}
}
\subfigure[\method (Gender)]{
\includegraphics[width=0.235\textwidth]{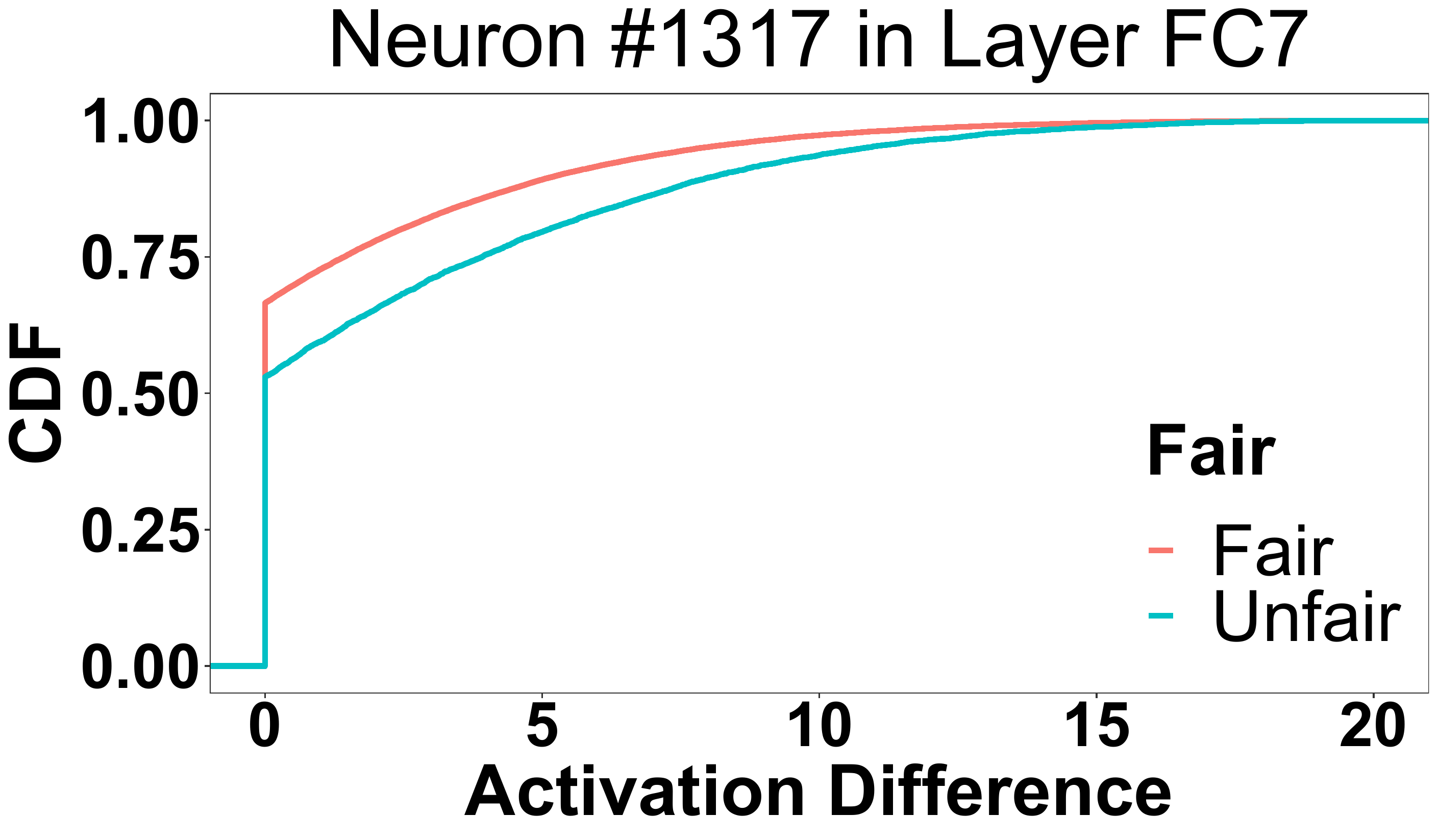}
\label{fig:gender}
}
\subfigure[DeepImportance (Race)]{
\includegraphics[width=0.235\textwidth]{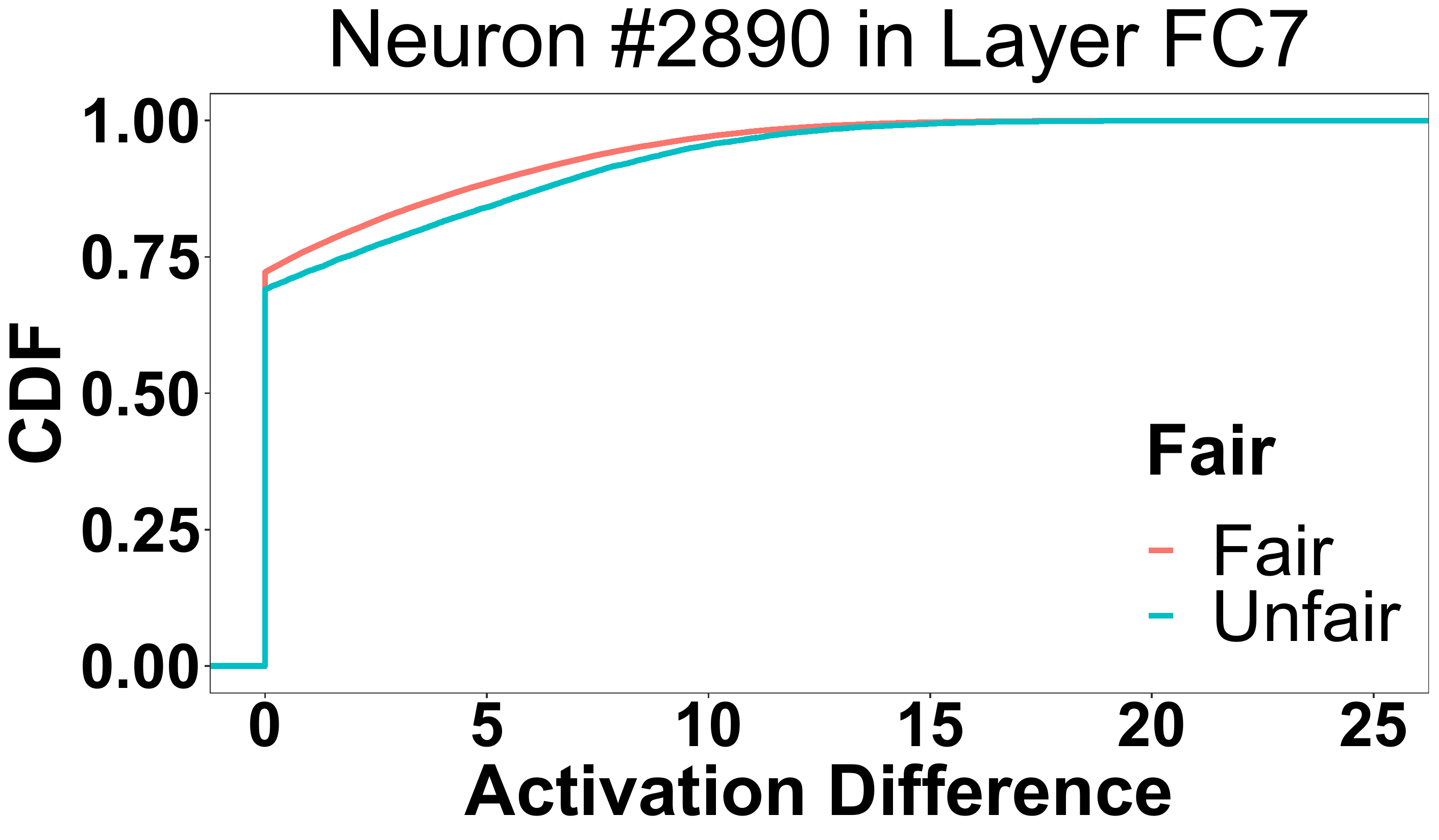}
\label{fig:idc}
}
\subfigure[Activation Importance (Race)]{
\includegraphics[width=0.235\textwidth]{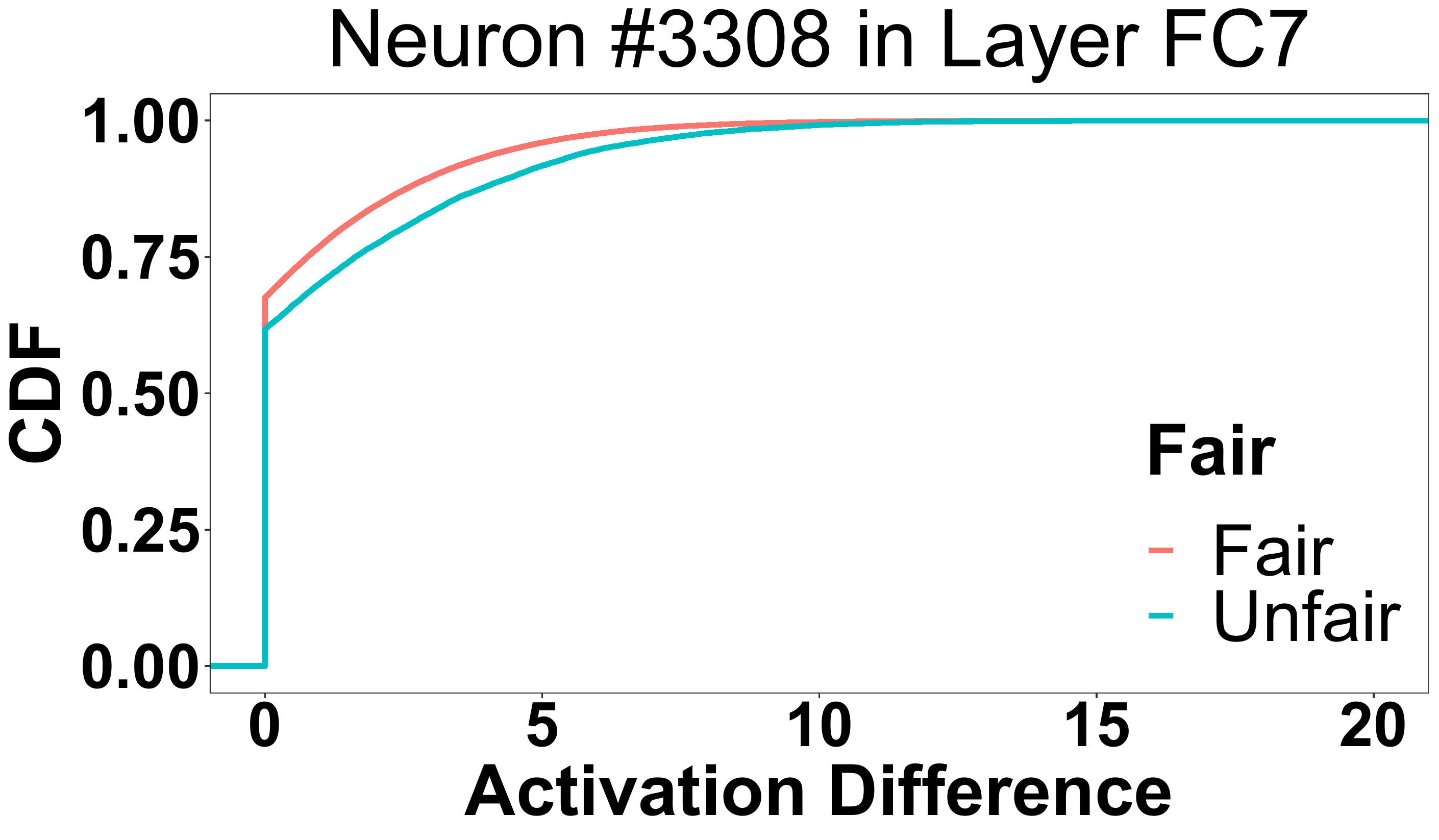}
\label{fig:av}
}
\caption{Cumulative distribution function (CDF) w.r.t. the difference of the neuron activation values.}
\label{fig:cdf}
\end{figure*}

\noindent Recall that fairness-related neuron selection is key to improve the efficiency and effectiveness of fairness adequacy testing. In the following, we first compare \method with two baselines. First is DeepImportance~\cite{importance_driven}, which is a recent work aiming to identify the most relevant neurons w.r.t. the model's decision. The other is activation importance (ActImp hereafter), i.e., which identifies neurons with the maximum activation value in a layer. Figure~\ref{fig:cdf} shows the cumulative distribution function (CDF) on the activation differences for VGG-16. Each figure represents the top neuron within the last hidden layer selected by \method with respect to race (shown in Figure~\ref{fig:race}) and gender (shown in Figure~\ref{fig:gender}), DeepImportance (shown in Figure~\ref{fig:idc}), and ActImp (shown in Figure~\ref{fig:av}) respectively. The larger the gap, the more relevant to the model's fairness the neuron is. We observe that compared with \methode, the neuron selected by DeepImportance and ActImp is less related with the model's fairness. This is because they only capture the neuron's contribution to the decision based on a single sample, which makes them less sensitive to the behavioral differences of a pair of samples. 
Moreover, it is evident that for different sensitive attributes, the selected neurons would be different as well.
For instance, the most relevant neurons in FC7 for race and gender are $\#3369$ and $\#1317$ with H value of $9,309$ and $646$, respectively.

\begin{table}[t]
\centering
\caption{The number of fairness-related neurons of the deepest $5$ layers.. FC, Conv, and Maxpool represent the Fully-Connected layer, Convolutional layer, and Maxpool layer, respectively. The layers are listed from deep to shallow.}
\label{tab:number}
\begin{tabular}{|c|c|c|c|c|}
\hline
Model & Layer & Total Neuron & \multicolumn{2}{|c|}{Fairness-related Neuron} \\
\cline{4-5}
& & & Race & Gender \\ 
\hline
\multirow{5}{*}{VGG} & FC7 & 4096 & 3471 & 3272\\
& FC6 & 4096 & 2881 & 2672\\
& Maxpool5 & 512 & 502 & 492\\
& Conv5\_3 & 512 & 504 & 490\\
& Conv5\_2 & 512 & 512 & 509\\
\hline
\multirow{5}{*}{ResNet} & Conv4.2\_3 & 2048 & 2028 & 2044 \\
& Conv4.2\_2 & 512 & 487 & 508\\
& Conv4.2\_1 & 512 & 484 & 507\\
& Conv4.1\_3 & 2048 & 1904 & 2036 \\
& Conv4.1\_2 & 512 & 427 & 509\\
\hline
\end{tabular}
\end{table}

\begin{figure}[t]
\centering
\subfigure[VGGFace (Race)]{
\includegraphics[width=0.22\textwidth]{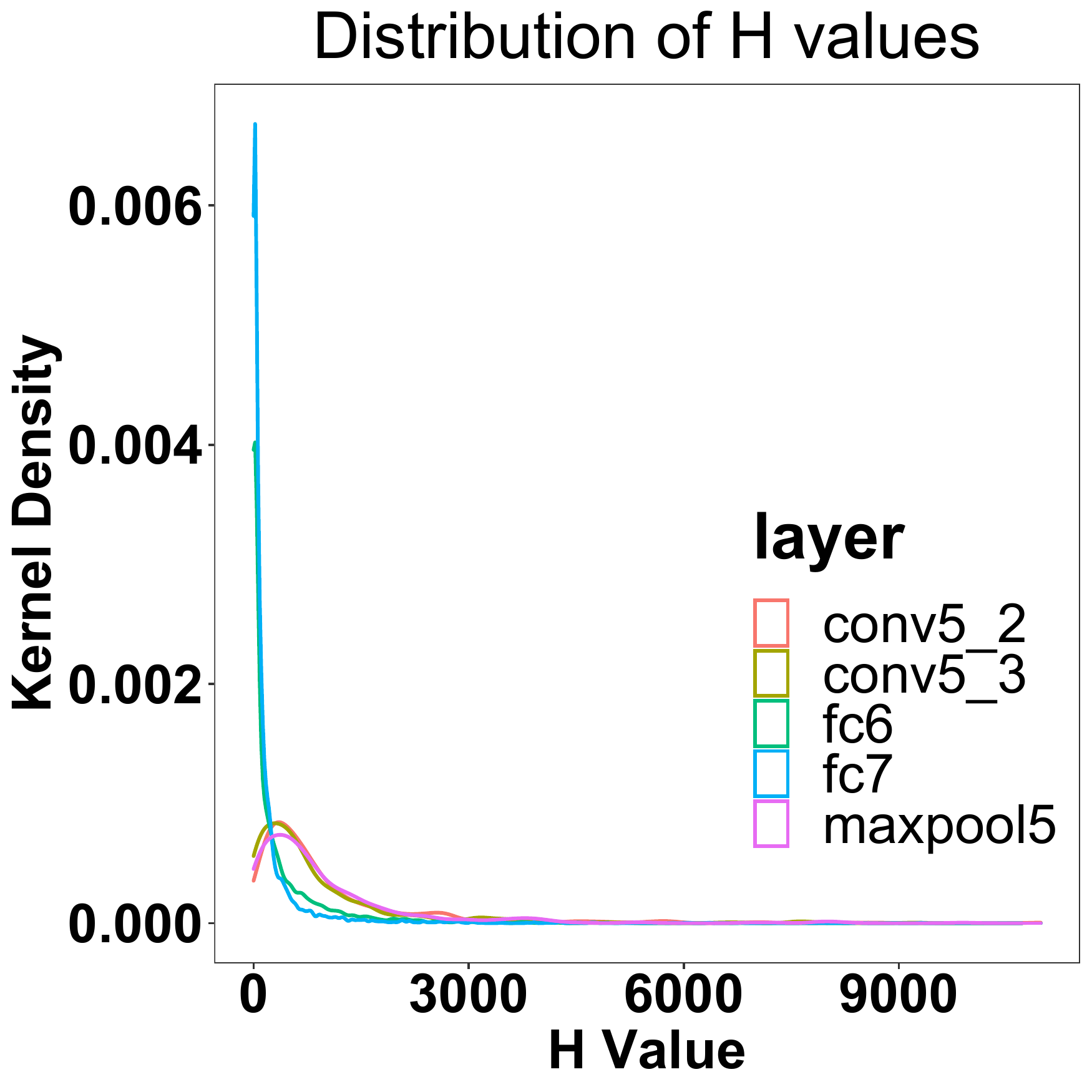}
\label{fig:h-race}
}
\subfigure[VGGFace (Gender)]{
\includegraphics[width=0.22\textwidth]{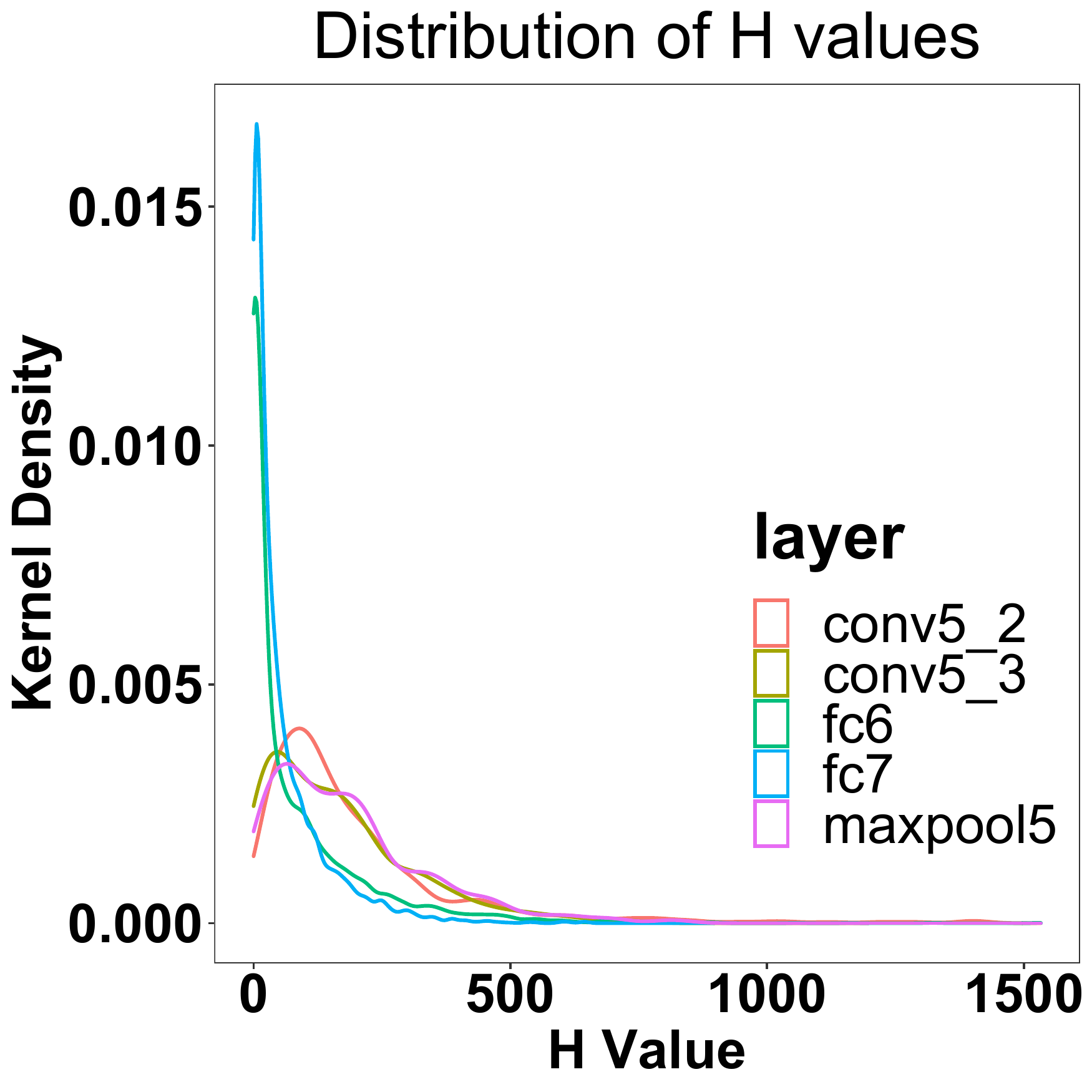}
\label{fig:h-gender}
}
\subfigure[FairFace (Race)]{
\includegraphics[width=0.22\textwidth]{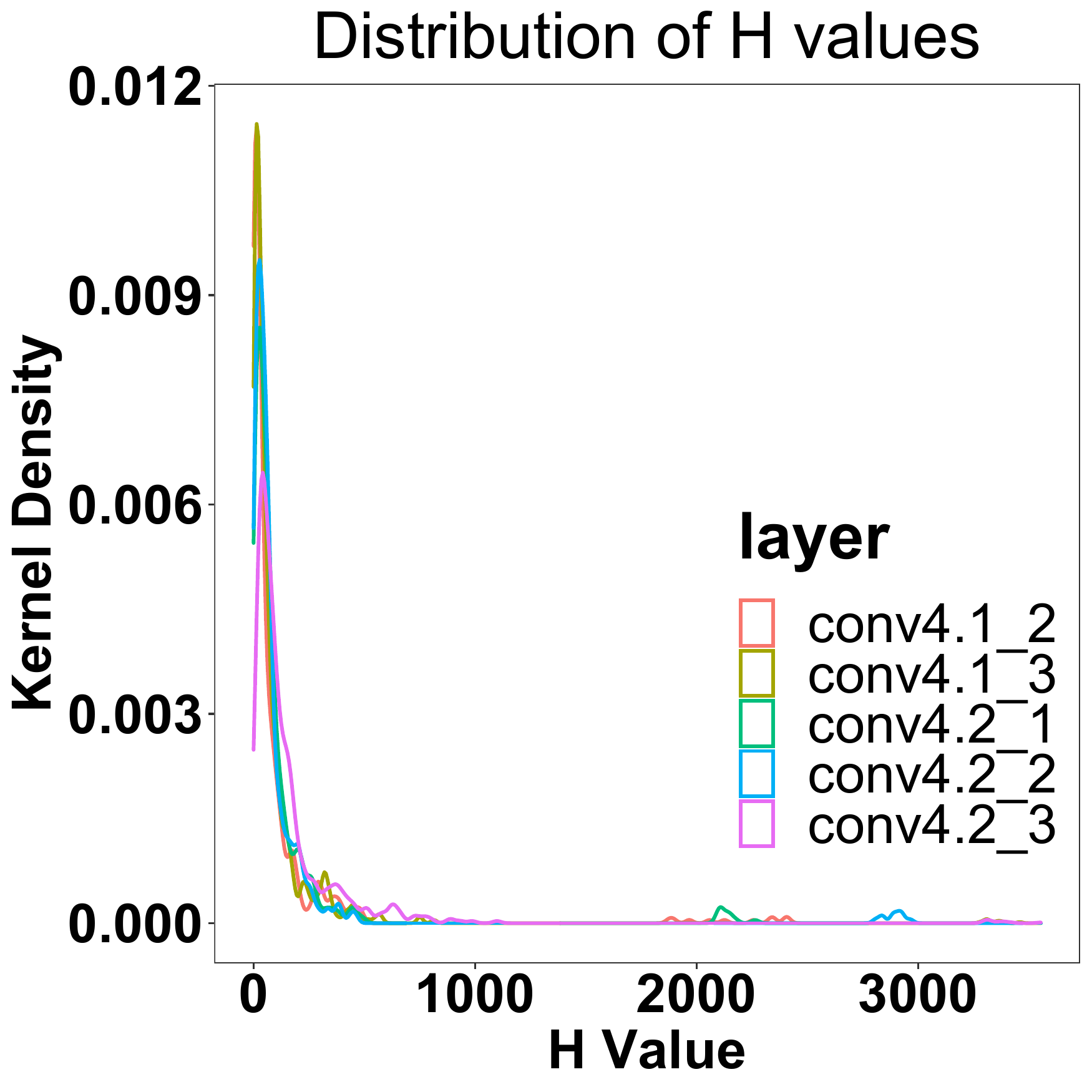}
\label{fig:h-race}
}
\subfigure[FairFace (Gender)]{
\includegraphics[width=0.22\textwidth]{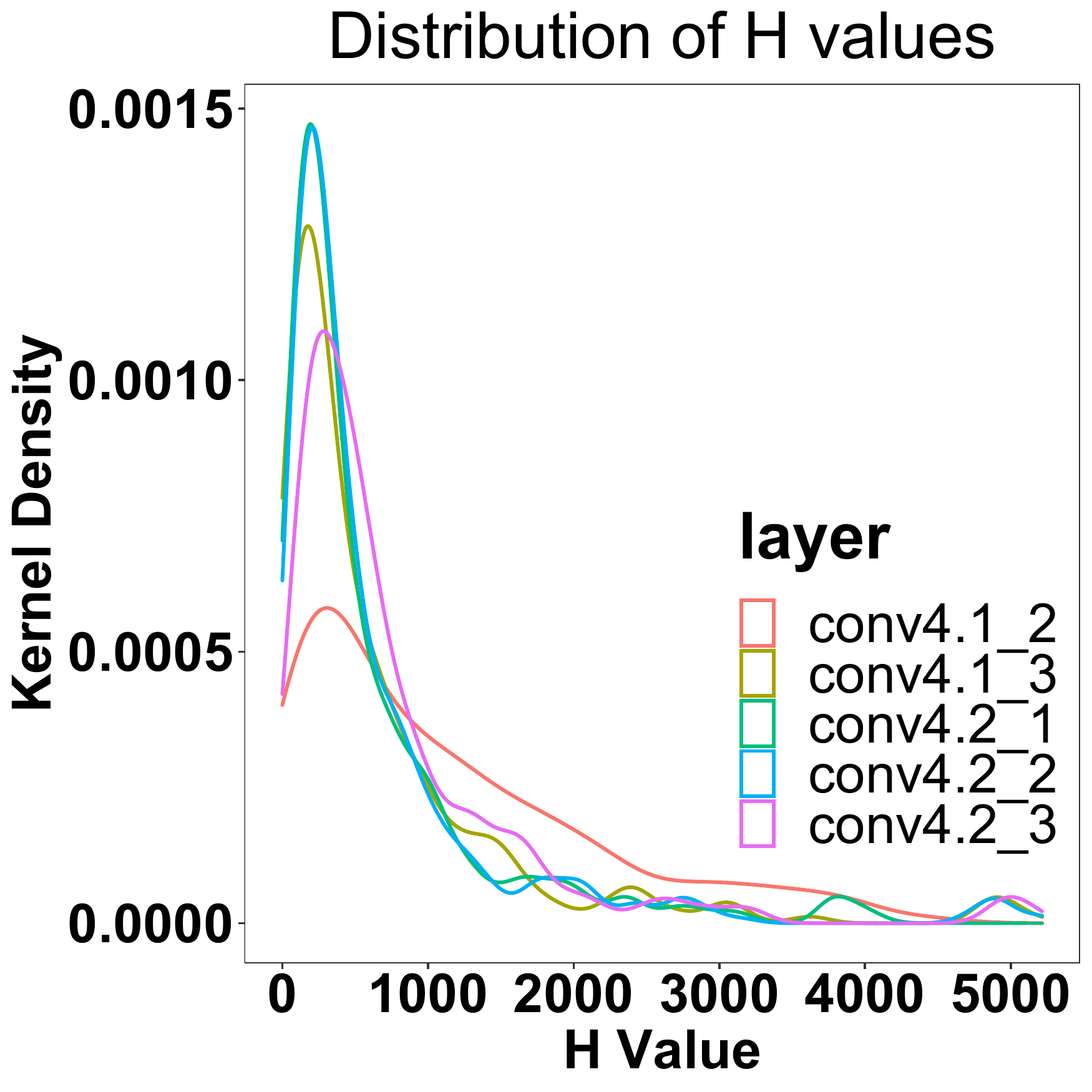}
\label{fig:h-gender}
}
\caption{The H distribution of the deepest $5$ layers.}
\label{fig:distribution}
\end{figure}

Table~\ref{tab:number} lists the number of total neurons and the selected fairness-related neurons of the deepest $5$ layers both on VGGFace and FairFace with respect to race and gender. We can observe that a considerable percentage of the neurons show correlation with fairness, for instance, the proportion is ranging from $70.34\%$ ($65.23\%$) to $100.00\%$ ($99.41\%$) on VGGFace with respect to race (gender). Further, Figure~\ref{fig:distribution} shows the H distribution of these $5$ layers. Note that the H values within each layer follows a long-tailed distribution, i.e., the H value of most neurons centralize on lower values, whereas a few values are pretty large, which shows significant correlation with fairness.

We thus have the following answer to RQ1,
\begin{framed}
\noindent \emph{Answer to RQ1: \method can effectively identify fairness-related neurons. While most of the neurons in the DNN are related to fairness, only a small percentage of them are strongly correlated.}
\end{framed}

\vspace{1mm}
\noindent\textbf{RQ2: How effective is \method for measuring the adequacy of fairness testing?}
\begin{table}[t]
\centering
\caption{Coverage performance ($\%$) w.r.t. layers.}
\label{tab:layer}
\begin{tabular}{|c|c||c|c|c||c|c|}
\hline
Layer & Data & Tan. & Cos. & Spe. & Abs. & Rel. \\
\hline
\multirow{3}{*}{FC7} & Fair & 69.50 & 61.00 & 34.30 & 31.04 & 37.54 \\
& Original & 78.30 & 78.60 & 42.50 & 40.73 & 49.21 \\
& Fair+GG & 86.20 & 87.10 & 45.90 & 37.41 & 47.78 \\
\hline
\multirow{3}{*}{FC6} & Fair & 57.60 & 46.10 & 29.80 & 41.44 & 61.00 \\
& Original & 65.00 & 59.70 & 36.70 & 49.31 & 68.52 \\
& Fair+GG & 72.30 & 67.40 & 40.50 & 47.62 & 68.46 \\
\hline
\multirow{3}{*}{Conv5\_3} & Fair & 16.40 & 36.70 & 29.20 & 43.83 & 78.92 \\
& Original & 17.60 & 44.80 & 34.50 & 50.86 & 82.45 \\
& Fair+GG & 17.60 & 49.10 & 37.40 & 49.95 & 83.84 \\
\hline
\multirow{3}{*}{Conv4\_2} & Fair & 0.20 & 13.70 & 25.90 & 26.40 & 45.89 \\
& Original & 0.20 & 14.80 & 27.20 & 29.35 & 46.01 \\
& Fair+GG & 0.20 & 15.10 & 28.40 & 27.76 & 46.00 \\
\hline
\end{tabular}
\end{table}

\noindent As presented in Section~\ref{subsec:metric}, all the criteria are calculated in a layer-wise manner. We thus first conduct an experiment to investigate the layer sensitivity of the testing adequacy before evaluating their effectiveness. 

We evaluate the testing coverage on $3$ data combinations. Table~\ref{tab:layer} shows the coverage computed on $4$ ordered layers, i.e., FC7, FC6, Conv5\_3, and Conv4\_2, of VGG-16 with respect to the sensitive attribute gender. Row \emph{Fair} and \emph{Ori.} are the coverage of $10,000$ fair pairs and adding all the discriminatory ones (since its size is smaller than $10,000$) of original dataset, and row \emph{Fair+GG} is the adequacy of the original fair data augmented with $10,000$ discriminatory instance pairs generated by the gradient-based strategy. 

First, it can be observed that for layer-level statistic, the coverage on the layers that are close to the output layer is significantly higher. For instance, the tanimoto coefficient of fair images drops from $69.50\%$ in FC7 to $57.60\%$ and $16.40\%$ in FC6 and Conv5\_3 and reaches $0.20\%$ in Conv4\_2. One possible explanation is that the deeper layers in the DNN are able to extract the more identity-related information. In addition, we also observe that: 1) a deeper layer is more sensitive to the original discriminatory samples, e.g., compared with the fair images, the absolute distance adequacy of the whole original testing dataset increases by $2.95\%$, $7.03\%$, $7.87\%$, and $9.69\%$ for Conv4\_2, Conv5\_3, FC6, and FC7 respectively; 2) the generated test cases show similar layer sensitivity, e.g., when we augment the original fair data with the one generated using the gradient-based strategy, the coverage of relative distance from FC7 to Conv4\_2 increases by $10.24\%$, $7.46\%$, $3.92\%$, and $0.11\%$, respectively. It is obvious that the selection of layers will affect the effectiveness of the testing. We thus suggest to choose the deeper layers to measure the adequacy of test cases. 

\begin{figure}[t]
\centering
\includegraphics[width=0.45\textwidth]{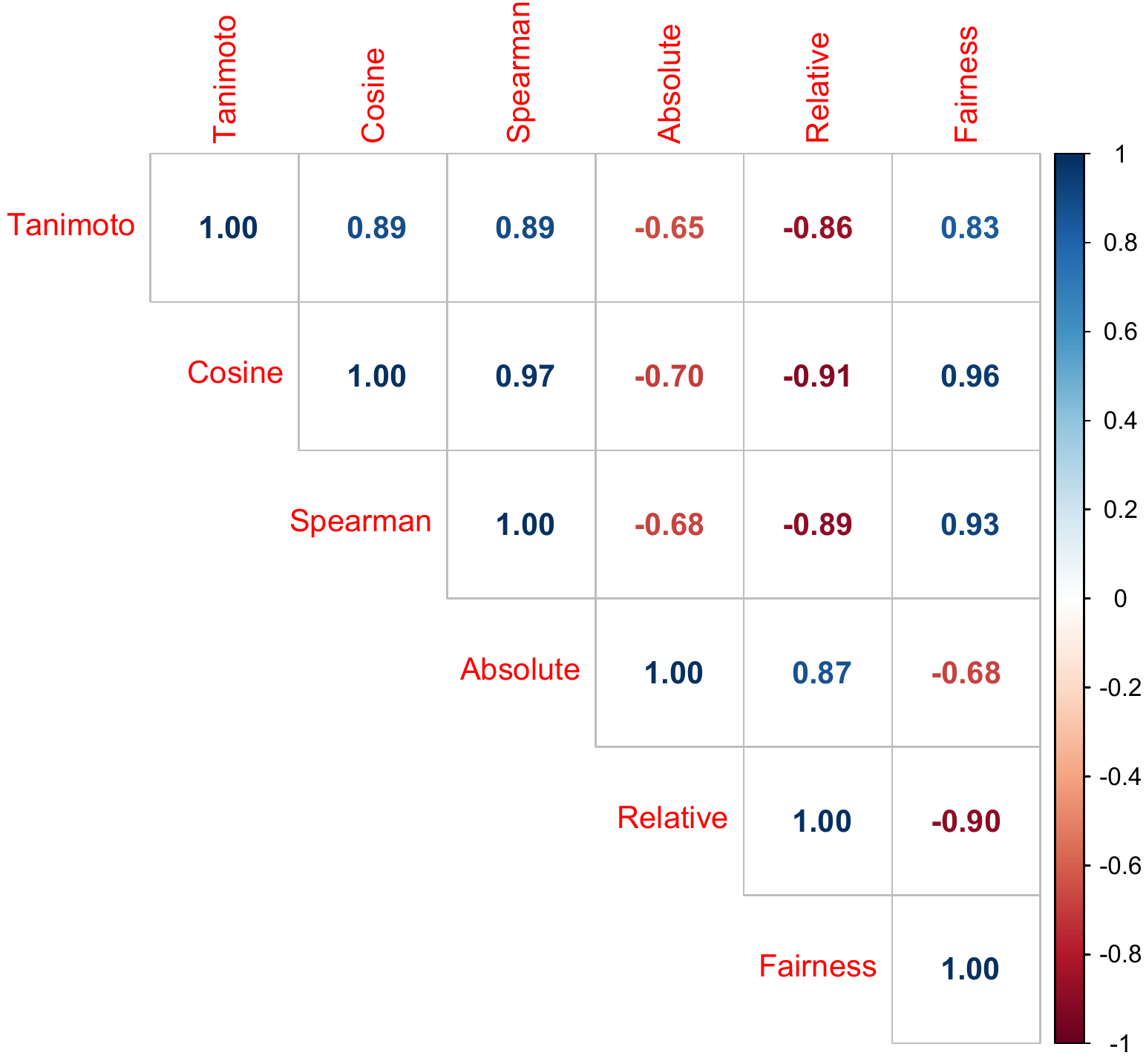}
\caption{Test coverage vs. fairness score.}
\label{fig:ccorrelation}
\end{figure}

Next, we conduct a correlation analysis on the coverage metrics and the individual fairness of the experimented models. We adopt the model mutation technique developed in~\cite{model_mutation,deep_mutation} to obtain a significant number of models with different behaviors efficiently for the study. In this work, we generate $10$ models for each mutation operator, e.g., Gaussian Fuzzing, Weight Shuffling, Neuron Switch, and Neuron Activation Inverse. Besides, since it is almost impossible to obtain meaningful images through random sampling in the input space, we measure the individual fairness of the model based on the ratio of non-discriminatory instances in the original testing dataset. The ranges of the accuracy and fairness scores of these $40$ models are in the range of $[94.00\%, 97.28\%]$ and $[64.37\%, 89.10\%]$, respectively. In order to ensure the fairness of the experiment, we randomly selected $10,000$ non-discriminatory image pairs and obtain the coverage on the deepest hidden layer, i.e., FC7.

We show the Pearson product-moment correlation~\cite{pearson} results on VGGFace with sensitive attribute gender in Figure~\ref{fig:ccorrelation}. The number in each cell is the correlation value between the metrics of the corresponding row and column, which ranges from $\-1$ to $1$. Note that all the correlation is significant, i.e., $p < 0.05$.

From Figure~\ref{fig:ccorrelation}, we have the following observations. First, the three layer-level criteria are significantly positively correlated with the fairness score, i.e., with a minimum correlation coefficient of $0.85$. It indicts that if the non-discriminatory instances has a higher coverage on the layer statistics, the DNN is fairer. This is intuitively expected, i.e., a fairer model can tolerate greater behavioral differences. Moreover, the two neuron distances show highly negative correlation with the individual fairness, i.e., with the value of $0.69$ and $0.90$. Our explanation is that for a fair model, the output difference of the neurons is small (so as to ensure that the final prediction does not change). Second, it is obvious that Tanimoto, cosine, and Spearman similarity have strong positive correlations with each other, while there is a moderate negative correlations between layer-level and neuron-level criteria.

\begin{table*}[t]
\centering
\caption{The adequacy of criteria on different data settings. }
\label{tab:effective}
{\small \begin{tabular}{|c|c|c||c||c|c|c|c||}
\hline
Dataset & Protected Attr. & Criteria & Fair & Fair+Unfair & Fair+GG & Fair+RG & Fair+GI \\
\hline
\multirow{12}{*}{VGGFace} & \multirow{6}{*}{Race}  & Tanimoto & 70.90 & 84.00 & 86.40 & 84.00 & 78.50\\
& & Cosine & 30.00 & 38.20 & 42.50 & 52.60 & 42.20\\
& & Spearman & 28.40 & 38.50 & 40.30 & 42.70 & 37.50\\
& & Abs. Distance & 30.88 & 46.66 & 49.68 & 48.01 & 44.44 \\
& & Rel. Distance & 31.10 & 42.50 & 53.54 & 53.34 & 46.65 \\
& & DeepImportance & 0.78 & 1.10 & 1.35 & 1.42 & 1.23 \\
\cline{2-8}
& \multirow{6}{*}{Gender}  & Tanimoto & 69.50 & 78.30 & 86.20 & 78.10 & 74.00\\
& & Cosine & 61.00 & 78.60 & 87.10 & 77.70 & 69.50 \\
& & Spearman & 34.30 & 42.50 & 45.90 & 41.30 & 37.90 \\
& & Abs. Distance & 31.04 & 40.73 & 37.41 & 39.27 & 39.87 \\
& & Rel. Distance & 37.54 & 49.21 & 47.78 & 51.34 & 52.68\\
& & DeepImportance & 0.93 & 1.28 & 1.26 & 1.38 & 1.35 \\
\hline
\multirow{12}{*}{FairFace} & \multirow{6}{*}{Race}  & Tanimoto & 20.30 & 22.50 & 35.90 & 39.00 & 24.80 \\
& & Cosine & 32.50 & 46.10 & 82.00 & 79.80 & 40.90 \\
& & Spearman & 20.10 & 23.20 & 45.30 & 38.80 & 22.00 \\
& & Abs. Distance & 23.42 & 31.13 & 58.49 & 41.55 & 28.18 \\
& & Rel. Distance & 39.94 & 53.37 & 78.01 & 58.48 & 50.83 \\
& & DeepImportance & 25.29 & 29.68 & 30.56 & 32.61 & 34.76 \\
\cline{2-8}
& \multirow{6}{*}{Gender} & Tanimoto & 20.60 & 23.70 & 34.20 & 32.90 & 23.90 \\
& & Cosine & 27.50 & 46.10 & 80.10 & 57.30 & 35.80 \\
& & Spearman & 17.70 & 23.30 & 43.40 & 28.90 & 20.70 \\
& & Abs. Distance & 32.31 & 45.43 & 81.85 & 37.57 & 37.17 \\
& & Rel. Distance & 39.85 & 55.80 & 68.47 & 50.77 & 47.29 \\
& & DeepImportance & 26.56 & 32.61 & 30.27 & 33.20 & 33.20 \\
\hline
\end{tabular}}
\end{table*}

In traditional software testing, coverage is not only a measure of testing adequacy, but also an effective tool for revealing bugs~\cite{survey} (i.e., by generating tests with high coverage). In this work, a bug refers to whether individual discriminatory samples exist in a given DNN model. Therefore, we conduct an effectiveness evaluation by comparing the coverage of testing criteria on the fair dataset and the dataset augmented with the individual discriminatory samples. The latter is obtained in two ways. One is the discrimination in the original testing data (column Fair+UnFair), and the other is discriminatory instances generated using $3$ different approaches, e.g., gradient-based generation (column Fair+GG), random generation (column Fair+RG), and Gaussian-noise injection (column Fair+GI). The parameters for generation are shown in Table~\ref{tab:configuration}.

For each dataset and each sensitive attribute, we random select $10,000$ image pairs for all the testing subsets including fair and discriminatory data (if the number is less than $10,000$, we take all of them) in the original set, and generated individual discriminatory pairs. The coverage result of the penultimate layer is presented in Table~\ref{tab:effective}. We repeat the procedure $5$ times and report the average coverage to avoid the effect of randomness. It can be observed that compared with the original fair pairs, the coverage of all the criteria has a significant increase after adding the discriminatory ones. Furthermore, compared with the discriminatory pairs in the original set and generated with image processing technology, fairness testing methods lead to higher coverage on all the criteria in most of the cases, except for the absolute distance and relative distance of VGG-16 with respect to gender. This is in line with our expectation, i.e., the criteria are more sensitive with individual discriminatory pairs generated by fairness testing, especially by the gradient-based method. In addition, we further conduct a comparison with DeepImportance~\cite{importance_driven}. It is worth noting that DeepImportance reports similar layer sensitivity in its evaluation. For a fair comparison, we also take the $10$ most important neurons, and the 
The total number of important neurons cluster combinations on VGG-16 and ResNet-50 are $131,072$ and $1,024$, respectively. We observe that the coverage of DeepImportance also increases when discriminatory samples are added. One possible reason is that the generation of unfair test case will push the seed towards the decision boundary~\cite{adf,tse}, which will also reduce the robustness of the seed. However, it is also observed that the DeepImportance is less sensitive than \method in most of the cases. In addition, since DeepImportance is only computed on individual samples, it cannot distinguish the similarity between each transformed sample and the original one, when sensitive attributes have multiple values.

We have the following answer to RQ2,
\begin{framed}
\noindent \emph{Answer to RQ2: The performance of our proposed fairness testing criteria vary across the layers, i.e., the deeper the layer, the more significant it is. Furthermore, they are strongly correlated with the fairness of DNN. Specifically, the layer-level statistics show positive correlation, whereas the neuron distances show negative correlation. The fairness testing criteria are effective for measuring the testing adequacy and capable for identifying individual discriminatory instances.}
\end{framed}

\vspace{1mm}
\noindent\textbf{RQ3: How effective are \method for test case selection?}
\begin{figure}[t]
\centering
\subfigure[Race]{
\includegraphics[width=0.225\textwidth]{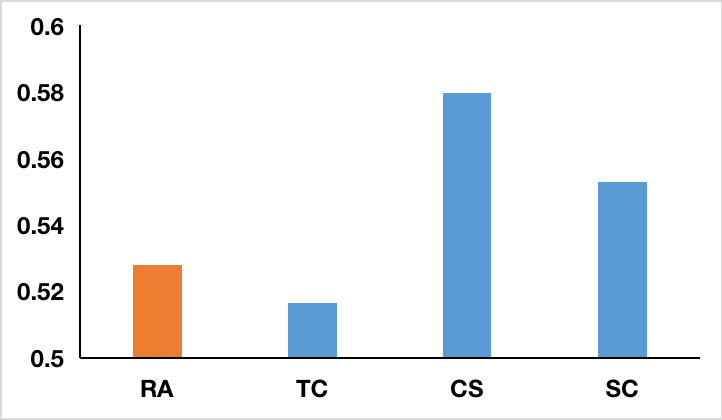}
\label{fig:im-race}
}
\subfigure[Gender]{
\includegraphics[width=0.225\textwidth]{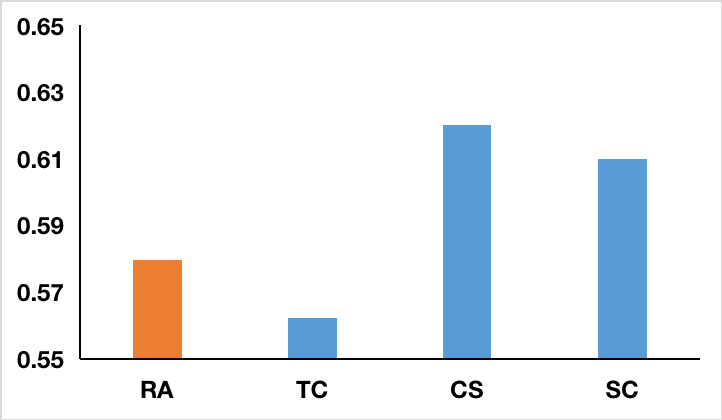}
\label{fig:im-gender}
}
\caption{Fairness improvement with different test case selection strageties.}
\label{fig:improvement}
\end{figure}

\noindent To answer this question, we evaluate the fairness improvement of augmented training, where the augmented data is selected by the KM-ST and completely random strategy (baseline), respectively. The training images are generated by the gradient-based method based on the original training data, and the validation set is composed by $5,000$ unfair samples each selected from the original testing set and three synthesized sets~\cite{robot}. Note that the training set and the validation set are disjoint. Recall that the neuron distance is a vector so that it is not a good quantitative metric for test case selection, and DeepImportance has the same problem. Thus we only apply KM-ST on the layer-level criteria.

Figure~\ref{fig:improvement} shows the results on dataset Fairface. Since we randomly select $10\%$ of the generated unfair samples for data augmentation, we present the average improvement of $5$ repeats. All models after augmented training have only a slight loss of accuracy, i.e., with a maximum decrease of $1.44\%$. From left to right, each bar represents the completely random strategy (RA) and KM-ST on Tanimoto Coefficient (TC), Cosine Similarity (CS), and Spearman Correlation (SC), respectively. We observe that compare with completely random: 1) applying KM-ST on cosine similarity and Spearman correlation are capable of reducing more discrimination in the model, especially applying it based on the cosine similarity, which has the greatest improvement of $5.15\%$ and $4.08\%$ with respect to race and gender, respectively. The reason is that compared with completely random method, KM-ST uniformly samples in each smaller subspace, which can better ensure that the obtained unfair samples are representative; 2) applying KM-ST on Tanimoto coefficient has $1.16\%$ and $1.73\%$ less fairness improvement with respect to race and gender, respectively. The reason is that compared with cosine similarity and Spearman coefficient, Tanimoto coefficient only considers the activation of neurons in each layer, while ignores some more fine-grained information such as the activation value.

We have the following answer to RQ3,
\begin{framed}
\noindent \emph{Answer to RQ3: Compared with completely random, applying KM-ST based on the proposed criteria of \method is more effective for selecting test cases to reduce the model's discrimination.}
\end{framed}

\subsection{Threat to Validity}
\label{subsec:discussion}

\noindent \textbf{Limited Subjects} Our experimental subjects (i.e., the datasets and DNN models) are limited. It might be interesting to conduct further evaluation on other datasets like VGGFace2~\cite{vggface2}, and other model structures like recurrent neural network (RNN)~\cite{lime}.

\vspace{1mm}
\noindent \textbf{Limitation of Domain Transfer} We adopt the image-to-image transformation approach, CycleGAN, for the transfer of sensitive facial attribute. The transforming process may have its limitations, such as translating the attributes which implicitly encode unique identity and the illumination variation. However, how to transfer images across sensitive domains while retaining as much other information as possible is still an open problem, and we will further investigate other possible directions.

\vspace{1mm}
\noindent \textbf{Limitation of Test Case Generation} We notice the other methods like coverage-guided fuzzing utilizing the proposed metrics, which we will further explore in future work.

\section{Related Work}
\label{sec:relatedwork}


\noindent \emph{\textbf{Fairness Testing}} Galhotra \emph{et al.}~\cite{themis} were the first to propose the fairness testing of machine learning, and utilized random generation to evaluate the frequency of discriminatory samples. Later, Udeshi \emph{et al.}~\cite{aequitas} improved it through using a two-step random strategy, AEQUITAS. The first stage is global generation, which samples the discriminatory cases in the input space at completely random. The second stage is local generation, it randomly searches the neighborhood of identified discrimination based on a dynamically updated probability. Besides, AEQUITAS tried to improve the model fairness through automatic augmentation retraining. Agarwal \emph{et al.}~\cite{sg} acquired unfair test cases by applying symbolic execution~\cite{concolic} to solve the unexplored path on local explanation tree~\cite{lime}. Its global and local generation aim to maximize the path coverage (diversity) and the number of instances respectively. Zhang \emph{et al.}~\cite{adf,tse} proposed a DNN-specific algorithm to search the discrimination based on gradient. It first maximize the output difference between input pair, and then perturb the identified one with less impact. The difference between \method and the above-mentioned works is two-fold. First, we introduce fairness testing to image data by domain transformation, instead of substituting the value of sensitive attribute with pre-defined one on tabular and text data. Second, previous works paid more attention to the generation of test cases, but ignored to measure the adequacy of testing.

\vspace{1mm}

\noindent \emph{\textbf{Robustness Testing Criteria}} Lots of robustness testing criteria were proposed. In~\cite{deepxplore}, Pei \emph{et al.} designed the first white-box robustness testing framework, DeepXplore, in the literature, which identifies and crafts unexpected instances with Neuron Coverage. In ~\cite{deepgauge}, Ma \emph{et al.} inherited the key insight and introduced five more fine-grained testing criteria both on layer and neuron levels.
In ~\cite{deepconcolic}, Sun \emph{et al.} brought the Modified Condition/Decision Cover into DL testing and proposed the first concolic testing method for DL models to improve four covering metrics based on a given neuron pair from adjacent layers, e.g., Sign-Sign Cover, Distance-Sign Cover, Sign-Value Cover, and Distance-Value Cover. In ~\cite{surprise}, Kim \emph{et al.} proposed two surprise coverage, LSC and DSC, which measures the range of the likelihood-based and distance-based adequacy values respectively. Later, ~\cite{iceccs,correlation} conducted the empirical study and showed that there is limited correlation between the robustness and the aforementioned coverage criteria of the DL model. In ~\cite{importance_driven}, Gerasimou \emph{et al.} calculate the contribution of each neuron to the final prediction through layer-wise backward propagation. In ~\cite{deepgini}, Feng \emph{et al.} proposed DeepGini, which prioritizes the unlabeled test cases by utilizing the impurity of output vector to reduce the resource consumption for labeling and better improve the robustness of model. More recently, Wang \emph{et al.} designed a robustness-oriented fuzzing framework (RobOT) based on the loss coverage (First-Order Loss)~\cite{robot}. Different from the above robustness testing metrics, our fairness testing criteria aim to measure the behavior difference between two similar samples, rather than the output of a single sample.

\vspace{-2mm}
\section{Conclusion}
\label{sec:conclusion}
In this paper, we bridge the gap in existing fairness testing research by proposing a novel testing framework, \methode, which systematically evaluates and improves the fairness testing adequacy of deep image classification applications. Our approach first selects the fairness-related neurons utilizing significance testing, then evaluates the fairness testing adequacy with five multi-granularity adequacy metrics and lastly selects the test cases based on the criteria for mitigating the discrimination efficiently. We evaluate \method on two large-scale public face recognition datasets. The results show that \method is effective both in identifying the fairness-related neurons, detecting unfair samples and selecting the representative test cases to improve the model's fairness.

\clearpage
\bibliographystyle{ACM-Reference-Format}
\bibliography{ft_criteria}
\end{document}